\documentclass[runningheads,dvipsnames]{llncs}
\usepackage{graphicx}
\usepackage{subcaption}
\usepackage{tikz}
\usepackage{comment}
\usepackage{amsmath,amssymb,bm}
\usepackage{color,soul}
\usepackage{xcolor}
\usepackage{multirow}
\usepackage{graphics}
\usepackage{enumitem}
\usepackage[space]{cite}
\usepackage[pagebackref,breaklinks,colorlinks]{hyperref}
\usepackage{bbding}

\hypersetup{
    colorlinks=true,
    linkcolor=black,
    citecolor=RoyalBlue,
    filecolor=magenta,
    urlcolor=magenta,
}

\begin{document}

\pagestyle{headings}
\mainmatter

\title{AutoTransition: Learning to Recommend Video Transition Effects}

\titlerunning{AutoTransition: Learning to Recommend Video Transition Effects}
\author{
    Yaojie Shen\inst{1,2,3}$^*$,
    Libo Zhang\inst{1,2}$^*$,
    Kai Xu\inst{3}$^*$,
    Xiaojie Jin\inst{3}$^{*\dagger}$
}
\authorrunning{Y. Shen et al.}
\institute{Institute of Software, Chinese Academy of Sciences \and
University of Chinese Academy of Sciences \and
ByteDance Inc.}

\def\thefootnote{*}\footnotetext{Equal contribution.}
\def\thefootnote{\textdagger}\footnotetext{Project lead and corresponding author $<$\url{jinxiaojie@bytedance.com}$>$.}
\def\thefootnote{\arabic{footnote}}

\maketitle

\begin{abstract}
    Video transition effects are widely used in video editing to connect shots for creating cohesive and visually appealing videos. However, it is challenging for non-professionals to choose best transitions due to the lack of cinematographic knowledge and design skills. In this paper, we present the premier work on performing automatic video transitions recommendation (VTR): given a sequence of raw video shots and companion audio, recommend video transitions for each pair of neighboring shots. To solve this task, we collect a large-scale video transition dataset using publicly available video templates on editing softwares. Then we formulate VTR as a multi-modal retrieval problem from vision/audio to video transitions and propose a novel multi-modal matching framework which consists of two parts. First we learn the embedding of video transitions through a video transition classification task. Then we propose a model to learn the matching correspondence from vision/audio inputs to video transitions. Specifically, the proposed model employs a multi-modal transformer to fuse vision and audio information, as well as capture the context cues in sequential transition outputs. Through both quantitative and qualitative experiments, we clearly demonstrate the effectiveness of our method. Notably, in the comprehensive user study, our method receives comparable scores compared with professional editors while improving the video editing efficiency by \textbf{300\scalebox{1.25}{$\times$}}. We hope our work serves to inspire other researchers to work on this new task. The dataset and codes are public at \url{https://github.com/acherstyx/AutoTransition}.  

\end{abstract}

\section{Introduction}

    With the advance of multimedia technology and network infrastructures, video is ubiquitous, occurring in numerous everyday activities such as education, entertainment, surveillance, etc. There is a massive amount of needs for people to edit videos and share with others. However, video editing is challenging for non-professionals since it is not only laborious but also needs a lot of cinematography and design knowledge. Some editing tools like \textit{Adobe Premier} and \textit{Apple Final Cut Pro} are developed to assist video editing, however their main target users are professionals while novices may find it difficult to learn. Moreover, they still lack the ability of automatic video editing, i.e., users have to manipulate videos on their own. Recently, popular video editing tools like \textit{InShot Video Editor} and \textit{CapCut} provide the function of creating videos in one-click. Nevertheless, since they only utilize simple strategies or fixed video templates and ignore the content of input vision/audio, the quality of generated video is unsatisfactory.
  
    The video transition effects play an important role in video editing to join shots for creating smooth and cohesive videos. In this paper, we introduce a new task of automatic video transition recommendation (VTR) and provide a systematic solution. Specifically, VTR is defined as: given a sequence of raw video shots and the companion audio (which can either be original sound or overwritten music), recommend a sequence of video transitions for each neighboring shots. Different from conventional classification problems which choose only one most probable category as the output, VTR aims to provide a ranking of candidate transition categories so that users can choose freely in practical usage.
    
    When working on this task, we encounter following challenges. 
    First, there is no publicly available video transition dataset for training and evaluation. It takes enormous efforts to manually collect and annotate a large-scale video dataset. Meanwhile, due to the large complexity and diversity of video editing, the evaluation of video quality is subjective and vary from person to person. Thus during creating the dataset, it is crucial to design proper criteria for selecting video samples that are appealing to most. 
    Besides dataset, solving VTR is also challenging. A good video transitions recommender should ensure top transitions match well with both the dynamics and contents of videos and the rhythm and theme of audio (or music). Moreover, transitions recommended at multiple video connections should be harmonious so that the final video is visually smooth and unified. Being the first effort in addressing VTR, we need to take all of above factors into consideration for delivering the optimal solution. 
    
    We start with building a video transition dataset from those video editing templates that are publicly available on video editing tools. 
    We design comprehensive rules via trials to select high-quality video templates followed by preprocessing for refinement. 
    Afterwards, we extract video shots and corresponding transitions (used as ground-truth in training), creating the first large-scale video transition dataset. More details are introduced in Sec. \ref{sec:data-collection}.
    
    To figure out the best way for modeling VTR, we conduct extensive experiments to compare the classification-based and retrieval-based solutions and finally demonstrate the latter performs better. In the classification-based solution, the model takes neighboring video shots as inputs and output the prediction of transition categories. The cross-entropy loss between predictions and ground-truth transitions is used.
    In the retrieval-based solution, we first pre-train a transition classification network to learn the embedding of transitions. Then we propose a multi-modal transformer to learn the fused vision/audio features in sequential video shots. A triplet margin loss is devised to minimize the distance between fused input features and pre-trained transition embedding. Similarly, in testing, according to the distance calculated between input features and transition embedding, the transition categories with smaller distances are in higher ranking position.
    
    \smallskip
    
    To conclude, our contributions are threefold: 
    \begin{enumerate}[leftmargin=*] {
        \item We introduce a new task of automatically recommending video transition effects given video and audio inputs. We collect the first large-scale video transition dataset to facilitate future research on this task.
        \item We formulate the transition recommendation task as a multi-modal retrieval problem and propose a framework for learning the correspondence between input vision/audio and video transitions in feature space. The proposed framework is capable of fully utilizing the multi-modal input information for generating sequential transition outputs.
        \item Through both quantitative and qualitative evaluation, we demonstrate that the proposed method can successfully learn the matching from vision/audio to transitions and generate reasonable recommendation results. Moreover, a user study conducted to evaluate the quality of generated videos further demonstrates the effectiveness of our method. 
        }
    \end{enumerate}

\section{Related work}

    \noindent\textbf{Video editing.}
    Automatic video editing is challenging due to following reasons.
    First, the model has to fully understand the spatial-temporal context and multi-modal information in videos to obtain semantically coherent results.
    Second, video editing requires lots of professional knowledge to endow videos with creativity and particular aesthetic taste. Third, the evaluation of the quality of the generated videos may be subjective.
    Recently, there are some progresses towards performing automatic video editing, each focusing on different aspects. Frey et al. \cite{frey2021automatic} proposes an automatic approach to transfer the editing styles of an edited video to the new raw shots. Wang et al. \cite{wang2019write} builds a tool for creating video montage based on the text description. Koorathota et al. \cite{koorathota2021editing} proposes a method to perform video editing according to a short text query by utilizing contextual and multi-modal information. Liao et al. \cite{liao2015audeosynth} introduces a method for music-driven video montage. 
    Several methods focus on solving video ordering and shot selection \cite{frey2021automatic,wang2019write,pardo2021learning}. Distinguished from all tasks above, VTR is still unexplored although its equal importance in creating high-quality videos in video editing. In this work, we take the first step to close the research gap.
    
    \smallskip
    \noindent\textbf{Video transitions.} Video transition is a widely used post-production technique for achieving smooth transitions between neighboring shots via special image/video transforms. There are various kinds of transitions including straight cuts, fades, and 3D animations among many others. 
    To professionals, each type of transition is with a dedicated meaning to convey specific emotions, feelings or scene information to viewers, thus should be used meticulously. Moreover, when multiple video transitions are used, they should work in a harmonious way to ensure the visual unification of the final video. Due to above reasons, it is difficult for non-professionals to apply video transitions in their edit. Our work can substantially assist these people by automatically recommending reasonable video transitions on the fly.
    
    \smallskip
    \noindent\textbf{Visual-semantic embedding.}
    Many recent works on video retrieval are based on the alignment of visual-semantic embedding \cite{hendricks2017localizing,akbari2021vatt}. Embedding techniques are employed to measure the similarity of different modalities in cross-modal video retrieval tasks, where features from different modals are mapped into a shared embedding space for better alignment \cite{karpathy2017deep,miech2019howto100m}. Miech et al. \cite{miech2019howto100m} utilizes millions of video text pairs to learn the text-video embedding. Escorcia et al. \cite{escorcia2019temporal} aligns the embedding of text and moments in the videos. They share the same idea of jointly aligning representations from two different modals. 
    Triplet loss is initially proposed for learning the distance metric \cite{schultz2003learning}. It is used for learning multi-modal embedding through deep neural networks in recent works \cite{faghri2017vse++,schroff2015facenet}. By optimizing the distance directly with a soft margin, triplet loss is suitable for ranking tasks~\cite{hoffer2015deep}.
    Our method also employs triplet loss to learn the distance between representations. A multi-modal transformer is used therein to learn the features of vision/audio inputs which aim to match with the pre-trained video transition embedding. Through extensive experiments, we demonstrate the effectiveness of our methods.
    
    \smallskip
    \noindent\textbf{Multi-modal transformer and sequence modeling.} 
    Our task is closely related to recent progress in modeling the spatial-temporal and multi-modal information in vision, speech and text. Transformer \cite{vaswani2017attention} is widely used in these tasks to encode cross-modal and spatial-temporal information \cite{arnab2021vivit,gabeur2020multi,lin2021survey,hendricks2021decoupling}.
    It employs an attention mechanism to represent multi-modal information in a common latent space. In other sequential problems, Lin et al. \cite{lin2021vx2text} uses a modality-specific classifier and a differentiable tokenization scheme to fuse multi-modal information via transformer. Gabeur et al. \cite{gabeur2020multi} introduces a video encoder architecture with multi-modal transformer for video retrieval. We also use modal-specific networks to extract embeddings from vision and audio inputs. Specifically, we use SlowFast \cite{feichtenhofer2019slowfast} and Harmonic CNN \cite{won2020data} to extract video and audio features respectively. 
    Different with the vanilla transformer architecture which adopts an encoder-decoder architecture~\cite{vaswani2017attention}, Lei et al.~\cite{lei2020mart} uses unified encoder-decoder transformer model for sequential modeling. 
    In our method, we exploit a multi-modal transformer for learning the multi-modal representations as well as capturing the context information in sequential transitions.  
  
    \begin{figure}[htbp]
        \centering
        \includegraphics[width=0.90\textwidth]{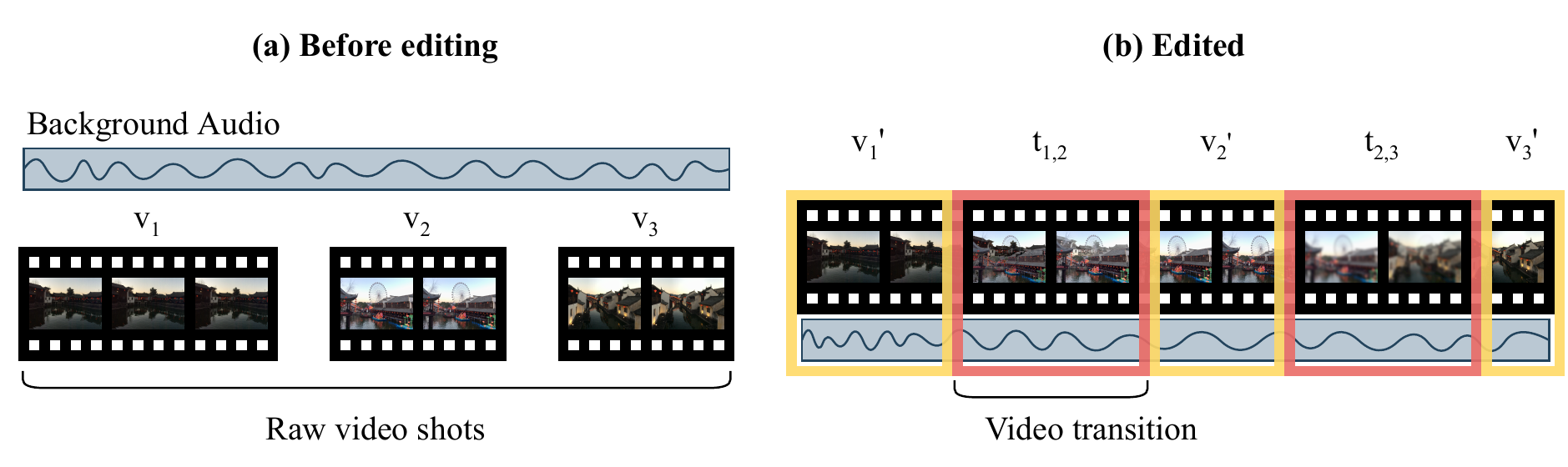}
        \caption{The definition of the task. We take a sequence of raw video shots, for example $\{v_1, v_2, v_3\}$, and companion audio as the inputs, the task is to recommend categories of the transitions $\{t_{1,2}, t_{2,3}\}$ used in the edited video.}
        \label{fig:task_defination}
    \end{figure}
        
    \section{Task Definition}
        
        We aim to solve the task of video transitions effect recommendation (VTR), the goal of which is to recommend appropriate transitions between neighboring video shots. As shown in Fig. \ref{fig:task_defination}a, we take a sequence of raw video shots $\{v_1, v_2, \dots , v_n\}$ as inputs. To simplify the task, we assume that the order of the videos is already determined, all the videos are already cut and scaled to the target range, and the background audio (either original sound and/or overwritten music) is already specified. Then for a pair of video shots $v_{k}$ and $v_{k+1}$, a transition effect is added to join them. We denote the video clip added with transition effect as $t_{k,k+1}$. 
        Since video transitions generally mix neighboring video shots, we separate the final video after adding transitions into two parts, one is the clips added by transitions (i.e. $t_{k,k+1}$), the other is uncontaminated video shots (i.e. $v_k'$ and $v_{k+1}'$).
        Then the output video after adding transitions can be denoted as $v'=\{v_1', t_{1,2}, v_2', \dots, t_{n-1,n}, v_n'\}$. The dataset we collect contains both the start and end timestamps of each transition so that we can obtain the exact positions of video shots and the categories of transitions. Note that in the collected dataset, we can only get access to the output video $v'$ since the original videos $v$ are not publicly available.
        
        In our method, we use the video clip $t_{k,k+1}$ and the label of corresponding transition effect $c_{k,k+1}$ to train the transition classification network, for learning the embedding of transitions based on their visual representation. When training the transition recommendation model, we remove $t_{k,k+1}$ from the input to be consistent with the inference setting where input videos are without any transition. The uncontaminated video shots $v_{k}'$ and $v_{k+1}'$ are used to represent the original video shots, i.e. $v_{k}$ and $v_{k+1}$ respectively. This strategy is reasonable since the duration of transition effect is short, $v_{k}'$ and $v_{k+1}'$ can serve as good approximations to their original counterparts.
        
    \section{Video Transition Dataset} \label{sec:data-collection}
    
        \subsection{Raw Data Collection}
        With the development of video editing tools and platforms, a large amount of well-designed video templates are publicly available. 
        Produced by professionals, video templates define fixed combinations of essential editing elements, including the number of video shots, the length of video, music, transitions, animations, and camera movement, etc. By simply replacing materials in templates with self videos, even novice users can create edited videos easily. 
        Each video template also comes with an example video made by the designer using this template and his/her original videos. In our experiments, we collect video templates from these online video platforms and  
        get the annotations related to transition effects, including each transition's
        category and corresponding start/end time. Though video templates may contain other special visual effects like animations and 3D movements, we only consider video transition effects in this paper.

        \subsection{Data Filtering}
        
        \begin{figure}[tbp]
            \centering
            \begin{subfigure}{0.46\textwidth}
              \includegraphics[width=\linewidth]{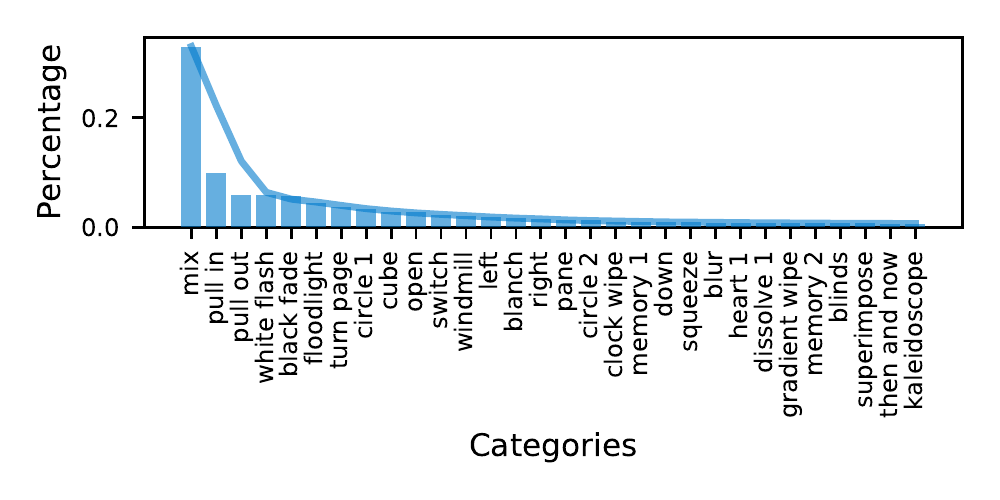}
              \caption{The label distribution on the small dataset at the first stage of data collection.}
              \label{fig:label_distribution_first}
            \end{subfigure}
            \hfill
            \begin{subfigure}{0.46\textwidth}
              \includegraphics[width=\linewidth]{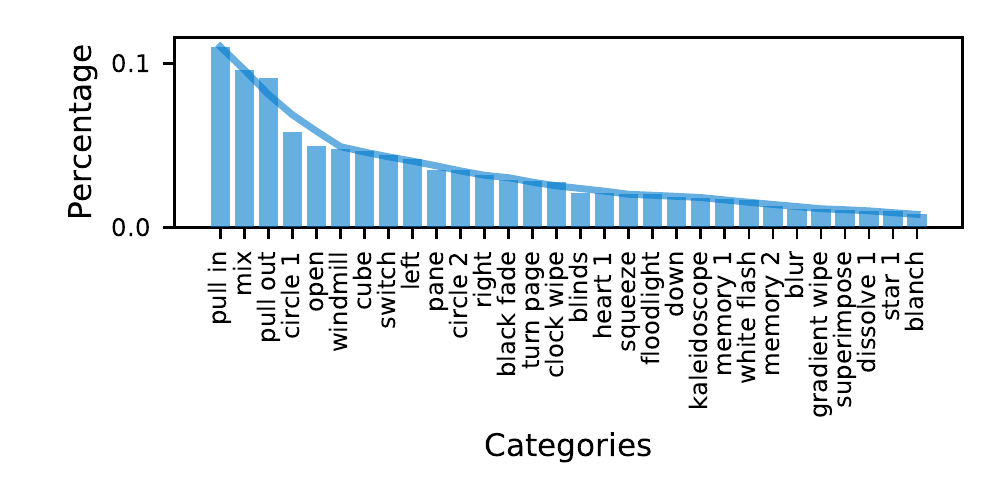}
              \caption{The label distribution on the final dataset after filtering.}
              \label{fig:label_distribution_second}
              \begin{minipage}{.1cm}
              \vfill
              \end{minipage}
            \end{subfigure}
            \caption{Label distribution of top-30 categories in the dataset.}
            \label{fig:label_distribution}
        \end{figure}
        
        We perform data filtering to improve the quality and diversity of the dataset.
        At first, we limit the maximum duration of the videos to 60 seconds, and the templates are filtered depending on the number of user likes and usages. We ignore those videos without any transition. We gather a small dataset through manual crawling and examine the overall distribution of collected samples. To guarantee that there are enough training samples for each category, we only select top-30 categories for training and testing according to the amount of samples. By statistical results, we observe that there are many duplicated transitions in a single video, and different types of transitions are distributed in a severe long-tail manner. The label distribution of the small dataset is shown in Fig. \ref{fig:label_distribution_first}.

        \setlength{\tabcolsep}{4pt}
        \begin{table}[tbp]
        \begin{center}
        \caption{The statistical information of the transition dataset.}
        \label{tab:dataset_statistic}
        \resizebox{0.5\textwidth}{!}{
        \begin{tabular}{c|cc}
            \hline
            Dataset & train set & test set \\
            \hline
            Number of videos & 29,998 & 5,000 \\
            Number of transitions & 118,984  & 19,869 \\
            Transitions per video & 3.966 & 3.973 \\
            Average video length & 15.83 secs  & 15.79 secs \\
            \hline
        \end{tabular}
        }
        \end{center}
        \end{table}
        \setlength{\tabcolsep}{1.4pt}
        
        We believe that the duplication and long-tailed distribution are harmful to the diversity of recommendation. To solve this issue, we use two additional rules to select samples: each video should contain more than two different types of transitions, and the usage times of the same transition should be no more than six. Following these rules, we acquire the final dataset which contains 34998 videos (train-29998 and test-5000) in total and 138.8K valid transitions between neighboring video shots. Table \ref{tab:dataset_statistic} shows more statistical results of the dataset. The label distribution is shown in Fig. \ref{fig:label_distribution_second}.

\section{Video Transition Recommendation}
    
    \subsection{Pre-training Transition Embedding} \label{section:pre-training_transition}
        
\begin{figure}[tbp]
        \centering
        \includegraphics[width=\textwidth]{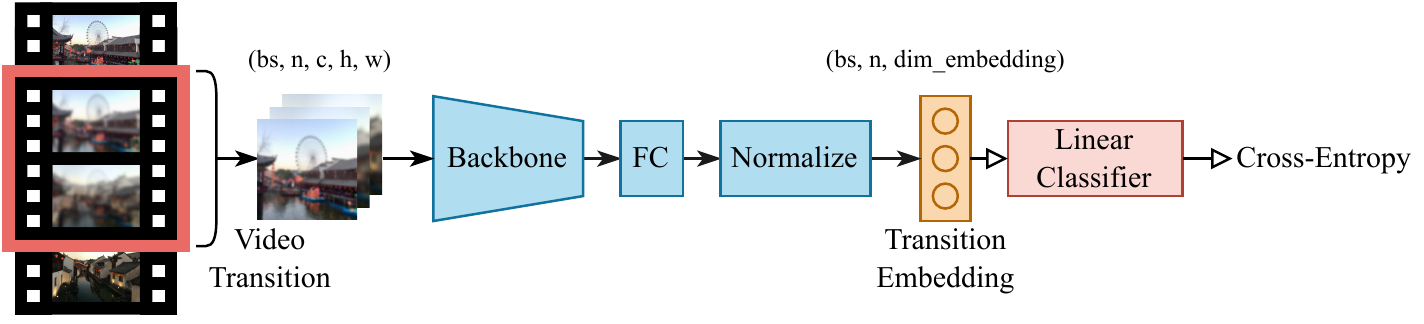}
        \caption{Extracting transition embedding. A transition classification network is built to learn the transition embedding.}
        \label{fig:pre-training_transition_embedding}
        \end{figure}

        We formulate the video transition effects recommendation as a multi-modal retrieval problem. Since retrieving from vision/audio to video transitions requires the model to learn correspondence between vision/audio inputs and video transitions, the first problem we need to solve is thus how to learn a strong representation for each transition. We notice that some video transitions have similar visual effects like ``pull in'' and ``pull out''. It is natural that we expect the learned embedding can also reflect these connections.
        To achieve this goal, we employ a video classification network to learn transition embedding based on their visual appearance. As shown in Fig. \ref{fig:pre-training_transition_embedding}, we take the transition clips $t$ as input and use the video backbone to extract visual representations. After passing through linear transform and normalization, we obtain a unit vector for each transition. Then we apply another linear transform and use the cross-entropy loss to optimize the classification objective. As expected, the embedding of the transitions are separably distributed in latent space and similar transitions stay close to each other. The visualization result of learned embedding through t-SNE is illustrated in Fig. \ref{fig:tsne}.
        
    \subsection{Multi-modal Transformer} \label{section:multi-modal trans}   
\begin{figure}[tbp]
            \centering
            \includegraphics[width=\textwidth]{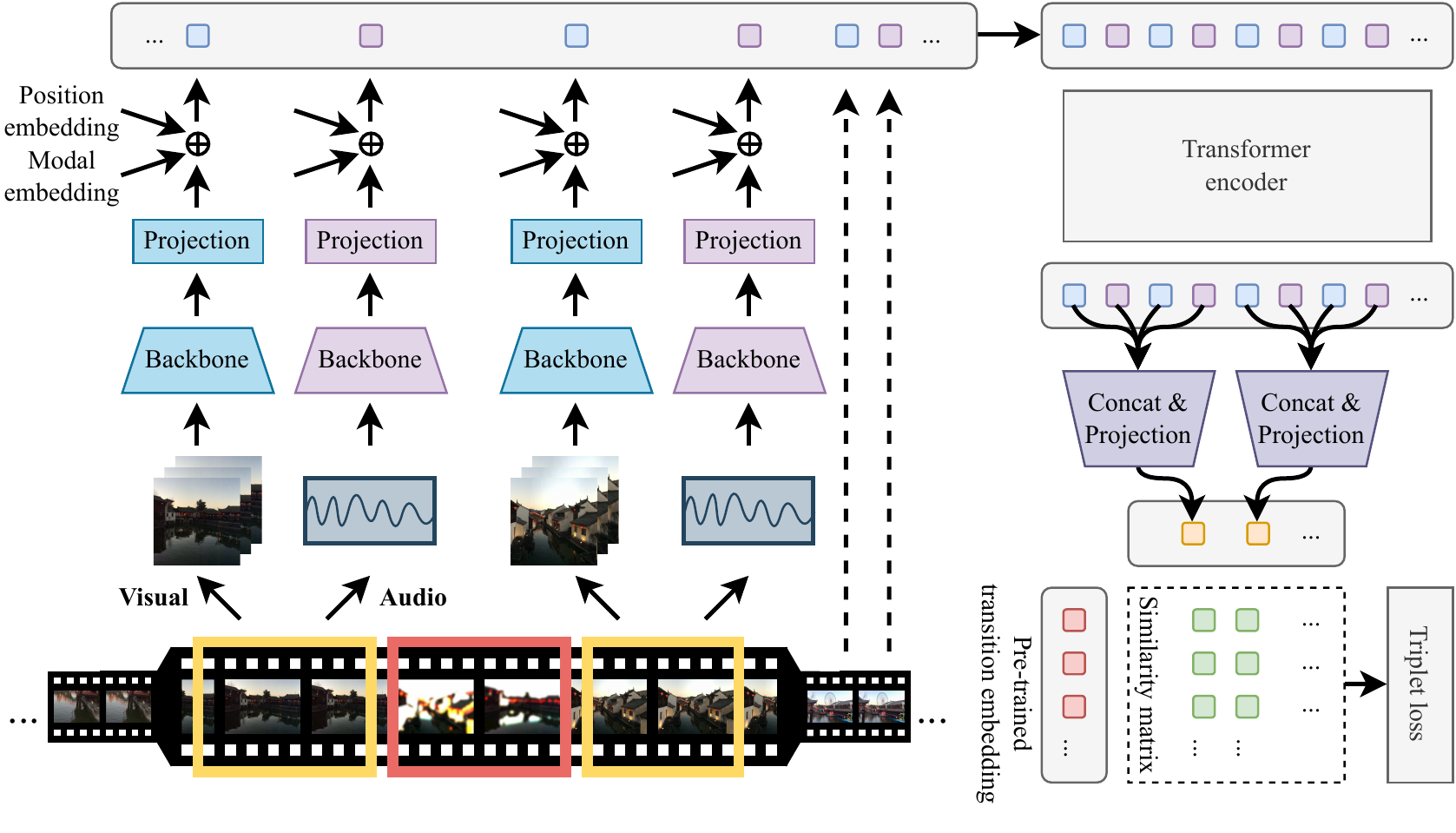}
            \caption{Transition recommendation model for retrieving matching transitions based on vision/audio inputs. First, we use modality-specific networks to extract visual and audio features. After that, a multi-modal transformer encoder fuses the tokens from different modalities. Finally, a triplet loss is used to optimize the network end-to-end. }
            \label{fig:framework_matching}
        \end{figure}

         As shown in Fig. \ref{fig:framework_matching}, we propose a multi-modal transformer to extract representations from the raw video shots and audio. For recommending a transition $t_{k,k+1}$ between video shots $v_{k}$ and $v_{k+1}$, we take both the video frames and audio waves from the end period of video shot $v_{k}$ and the start period of video shot $v_{k+1}$ as input. Specifically, $n$ video frames are sampled uniformly from each video shot.
         
         After obtaining video frames and corresponding audio waves, we extract their features by feeding into visual and audio backbone respectively. Note that these backbones can be conveniently replaced by other common video or audio models. In our experiments, we use the SlowFast \cite{feichtenhofer2019slowfast} and the Harmonic CNN \cite{won2020data} as video and audio backbone respectively.
         
         In order to make full use of multi-modal information of vision/audio, we combine the visual and audio features as the inputs for the multi-modal transformer. By doing so, the model not only learns the matching relationship from inputs to transitions but also captures the context cues among sequential transitions.
         
         Before being fed into the transformer, visual tokens and audio tokens are projected to the same dimension by independent linear transformations. Then learnable positional embedding and modal embedding are element-wisely added to these tokens. We share the positional embedding for vision/audio tokens at the same time point. Modal embedding is applied to inform the model which modality the token belongs to. After above processing, tokens from all video shots are input into transformer as a whole. In this way, the transformer is encouraged to learn the contextual relationship in sequential transitions. As demonstrated by experimental results, such a context-aware training method contributes to generate harmonious sequential outputs.
         The self-attention mechanism in transformer encoder can model complex mutual relationship among input tokens. From all output tokens, we concatenate those four tokens which corresponds to each transition point (for each modality of vision and audio, there are two tokens before and after the transition point) along the feature dimension and get the final representation by a projection to the same dimension of transition embedding.
        
    \subsection{Transition Recommendation}
    
        As introduced in Sec.~\ref{section:multi-modal trans}, the multi-modal video embedding is extracted by the multi-modal transformer, which is used as the query to retrieve the transitions by matching the pre-trained transition embedding mentioned in Section \ref{section:pre-training_transition}. During retrieval, we apply learnable linear transformations to both the pre-trained transition embedding and the video embedding for achieving better alignment. 
        
        We denote $E^{trans}= \{e^{trans}_1, \dots, e^{trans}_N\}$ as the set of pre-trained transition embedding where $N$ is the number of categories and $N=30$ in our experiments. $E^{video}= \{ e_1^{video}, \dots, e_{V}^{video} \} $ is denoted as the multi-modal input embedding in a batch. $V$ is the number of all transition points in a batch. For each sample $e_{v}^{video}$, we calculate its matching score with every transition embedding through a similarity metric $\mathrm{\Phi}(e_{v}^{video}, e^{trans}_k)$. In our implementation, $\mathrm{\Phi}$ adopts the form of dot-production due to its simplicity, i.e.
        \begin{equation}
        \label{eq:phi}
        \mathrm{\Phi}(e^{video}_v, e^{trans}_k) = \left<e_v^{video},e^{trans}_k \right>.            
        \end{equation}
        Eq.~(\ref{eq:phi}) is used to calculate the ranking loss in the following training steps.
        
        \smallskip
        \noindent\textbf{Training.}
        We expect that the model can learn to rank transitions by their distances with inputs in embedding space. To achieve so, we utilize the triplet margin loss to optimize the similarity between the transition embedding and multi-modal video embedding. For the embedding of each training sample $e^{video}$ with ground truth label $c$, we define our training objective with triplet loss as
        \begin{equation*}
        \mathcal{L}(e^{video}) = \frac{1}{N-1}\sum_{k\ne c,k\in \{1,\dots,N\}}{\mathcal{T}(e^{video},e^{trans}_c,e^{trans}_k)}            
        \end{equation*}
        where $\mathcal{T}$ calculates the triplet margin loss for each triplet $(e^{video}, e_c^{trans}, e_k^{trans})$:
        $$ \mathcal{T}(a,p,n) = max(\mathrm{\Phi}(a, p) - \mathrm{\Phi}(a, n) + M, 0). $$ $M$ is the soft margin, $a$, $p$ and, $n$ are anchor, positive sample, and negative sample, respectively. $\mathrm{\Phi}$ follows the definition of Eq.~(\ref{eq:phi}). In our settings, we take video embedding $e^{video}$ as the anchor, the transition embedding with category $c$ as the positive sample, others transitions as negative samples. The final objective is the average over all samples, i.e.
        \begin{equation}
        \label{eq:final-loss}
            \mathcal{L}(E^{video}) = \frac{1}{V}\sum_{v\in\{1,\dots,V\}}{\mathcal{L}(e^{video}_v)}.
        \end{equation}
        By optimizing Eq.~(\ref{eq:final-loss}), the model encourages the similarity between the multi-modal video embedding and its ground-truth transition embedding higher than the similarity between non-matching pairs with a margin of $M$. 
        
        \smallskip
        
        \noindent\textbf{Evaluation.}
        In evaluation, we follow Eq.~(\ref{eq:phi}) to measure the matching degree between multi-modal video embedding and candidate transition embedding. For two neighboring video shots, we sort the similarities of candidate transitions in descending order and select the top one as final result.
        
\section{Experiments}
    
    \subsection{Implementation Details}
        
        \noindent\textbf{Model details.} 
        We employ SlowFast 8$\times$8 \cite{feichtenhofer2019slowfast} as the backbone to extract visual features. The same backbone is also used as the transition embedding network. By default, we freeze SlowFast from stage 1 to stage 3 during our experiments to save memory. We train all models on one machine with 8 NVIDIA V100 GPUs, except the experiment without freezing the SlowFast backbone, in which we use two machines and 16 GPUs in total. For audio modal, we use Harmonic CNN \cite{won2020data} to extract the local audio features around the transition. 
        Then we linearly project the feature of two modals to the same dimension of $d_{model}$ and take them as the input tokens of the multi-modal transformer. The multi-modal transformer consists of two transformer layers with $d_{model} = 2048$ and $n_{head}=8$. 
        Before matching, we apply linear projections to both the video embedding and pre-trained transition embedding in order to map them into a joint space. Such a projection is experimentally proved to be beneficial as shown in Table \ref{tab:pre-training_transition_embedding}. 
        
        \smallskip

        \noindent\textbf{Data preprocessing and training.} 
        When training the transition embedding network, we uniformly sample 16 frames with the image size of 224$\times$224 from the transition duration as the input. The batch size is set to 256. The training process is 30 epochs in total, and start with a warm up for 5 epoch to raise the learning rate from 1e-6 to the initial learning rate 1e-3, then decay by a factor of 0.1 every 10 epochs. We use the model parameters of the last epoch to generate the transition embedding.

        When training the transition recommendation model, we uniformly sample 16 frames with the image size of 224$\times$224 as the visual inputs. The local audio features are extracted using the pre-trained Harmonic CNN before training. Given a time point, the Harmonic CNN generate a feature vector with a dimension of 100 based on the audio in around one second. For the sequential inputs, we set the maximum sequence length to 8. Redundant transitions beyond the maximum length are dropped while zero tensors are padded if the length is less than 8. We use Adam optimizer in all the experiments and set the soft margin $M=0.3$ in triplet margin loss. The initial learning rate is set to 1e-5, then decay by a factor of 0.1 every 10 epochs. The total training epoch is 30 epochs.

        \smallskip
        
        \noindent\textbf{Metrics.} For testing the fine-grained performance of our model, we evaluate the individual recommendation results in testing.
        The commonly used Recall@K metric ($k\in \{1,5\}$) and Mean Rank are employed as evaluation metrics. 
        Since in our dataset, there is only one ground-truth for each transition, Recall@K indicates the likelihood of hitting the target in top K retrieval results. 
        Mean rank is the averaged rank of all ground-truth transitions, whose math formula is
        $$
        \mathrm{MR} = \frac{1}{|\mathcal{S}|} \sum_{i \in {S}} \mathrm{Rank}(i)
        $$
        where $\mathcal{S}$ represents the set of all transitions in test.
        
    \subsection{Extracting Transition Embedding}
        
        \begin{figure}[tbp]
        \centering
        \includegraphics[width=0.40\textwidth]{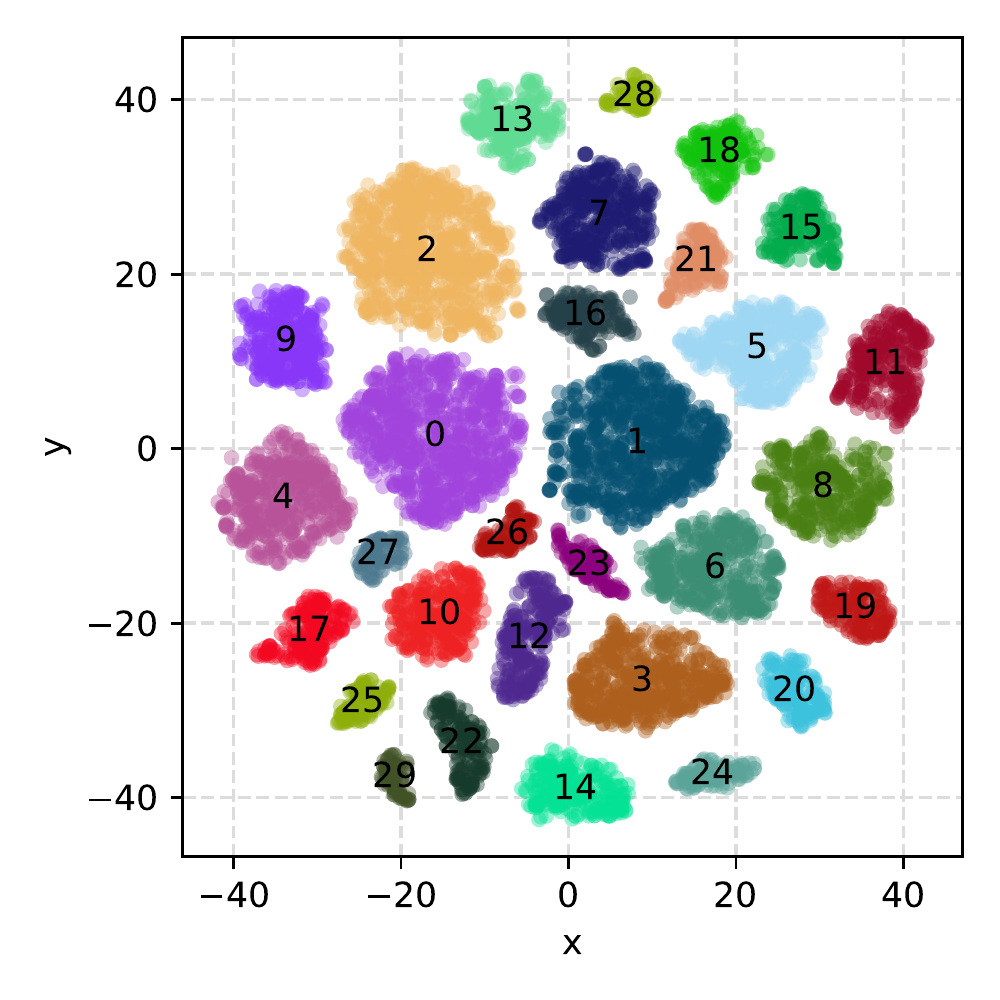}
        \includegraphics[width=0.41\textwidth]{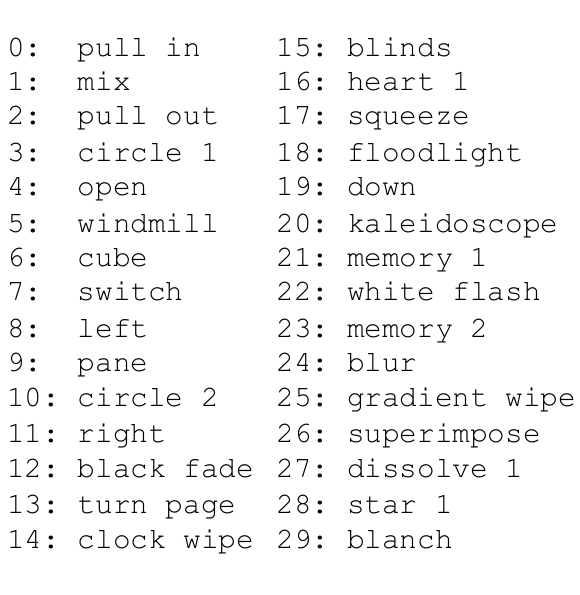}
        \caption{t-SNE visualization of the transition embedding on the train set after pre-training. We use cosine similarity as the distance metrics when running t-SNE. We drop the outliers and randomly pick 10K samples from the t-SNE output for visualization. We can see that transitions with similar visual effects, such as ``left'' (8), ``right'' (11), and ``down'' (19) are close to each other in the embedding space.}
        \label{fig:tsne}
        \end{figure}
        
        A transition classification network is trained using annotated transition category in the dataset as the transition embedding network. After training, we use the pre-trained network to extract the embeddings of
        all transitions in the training set. Normalization is then applied for the embedding as shown in Fig. \ref{fig:pre-training_transition_embedding} to convert the embedding into a unit vector. We drop the outliers utilizing the three-sigma rule by assuming the embeddings follow a normal distribution. The embeddings after dropping are visualized in Fig. \ref{fig:tsne}. The remaining embeddings for each category are averaged to generate the final transition embedding.
        We then demonstrate the effectiveness of the transition embedding by the experiments shown in Table \ref{tab:pre-training_transition_embedding}. Notably, the transition embedding network is advantageous in extending the model to support new transition categories since retraining the recommendation model is circumvented.

    \subsection{Ablation Studies and Comparisons}
        
        We start by showing the advantages of leveraging contextual and multi-modal information in inputs. Then we verify the effectiveness of the pre-trained transition embedding by comparing it with random initialization. We also compare with the classification method to demonstrate the superiority of retrieval methods on this task. 
        Due to space limit, more details of results are referred to the supplementary material.
        
        \smallskip
        \noindent\textbf{Context and multi-modal.} 
        In this experiment, we study the impact of contextual and multi-modal inputs on the recommendation performance, and the result is shown in Table \ref{tab:context_and_modal}. From the first and third rows of Table \ref{tab:context_and_modal}, we can see that sequential inputs introduce the context information to the model, which is helpful for modeling the temporal relations between the transitions. From the second and third rows of Table \ref{tab:context_and_modal}, visual modal inputs perform much better than audio as input, which indicates that visual content is more related to the transition effects than audio. 
        The results in the last three rows demonstrate that the multi-modal inputs can improve the accuracy of recommendations than the single modal inputs.
        
        \setlength{\tabcolsep}{4pt}
        \begin{table}[tbp]
        \begin{center}
            \caption{The impact of contextual and multi-modal information to the recommendation results.}
            \label{tab:context_and_modal}
            \begin{tabular}{c|cc|ccc}
                \hline
                % head
                \multirow{2}{*}{Sequential (Context)} & \multicolumn{2}{|c|}{Modal} & \multirow{2}{*}{Recall@1} & \multirow{2}{*}{Recall@5}  & \multirow{2}{*}{Mean Rank} \\
                \cline{2-3}
                
                & Visual & Audio & & & \\
                \hline
                % data
                                        & \checkmark&           & 24.12\% & 66.25\% & 5.758 \\
                \checkmark              &           & \checkmark& 19.39\% & 56.61\% & 7.012 \\
                \checkmark              & \checkmark&           & 25.40\% & 66.33\% & 5.665 \\
                \checkmark              & \checkmark& \checkmark& \textbf{28.06\%} & \textbf{66.85\%} & \textbf{5.480} \\
                \hline
            \end{tabular}
        \end{center}
        \end{table}
        \setlength{\tabcolsep}{1.4pt}
        
        \smallskip
        \noindent\textbf{The effectiveness of pre-trained transition embedding.}
        In this experiment, we study the effectiveness of our proposed pre-trained transition embedding, and the results are shown in Table \ref{tab:pre-training_transition_embedding}. All the embedding is frozen during training. In the random initialization setting, we use a normalized random embedding as the replacement of the pre-trained transition embeddings. It is observed that the performance of using random embedding is worse compared to our pre-trained embedding.
        From the results in second and third rows in Table \ref{tab:pre-training_transition_embedding}, the importance of the linear projection can be demonstrated. We conjecture the linear projection helps learning better mapping between the pre-trained transition embedding and multi-modal input embedding in a shared space.
        
        \setlength{\tabcolsep}{4pt}
        \begin{table}[tbp]
            \caption{Advantages of pre-training transition embedding from the visual effects. We demonstrate its effectiveness by comparing it with a random initialized embedding.}
            \label{tab:pre-training_transition_embedding}
            \centering
            \resizebox{\textwidth}{!}{
            \begin{tabular}{l|c|ccc}
                \hline
                Transition Embedding            & Projection    & Recall@1 & Recall@5 & Mean Rank \\
                \hline
                Random initialization               &               & 25.67\% & 66.3\% & 5.646 \\
                Pre-trained transition embedding &               & 26.24\% & 66.03\%  & 5.623 \\
                Pre-trained transition embedding & \checkmark    & \textbf{28.06\%} & \textbf{66.85\%} & \textbf{5.480}  \\
                \hline
            \end{tabular}
            }
        \end{table}
        \setlength{\tabcolsep}{1.4pt}
        
        \setlength{\tabcolsep}{4pt}
        \begin{table}[tp]
            \caption{Comparing with the classification method.}
            \label{tab:classification}
            \centering
            \resizebox{\textwidth}{!}{
            \begin{tabular}{l|ccc}
                \hline
                % head
                Methods       & Recall@1 & Recall@5 & Mean Rank \\
                \hline
                % data
                Classification                                  & 22.27\% & 61.82\% & 6.099 \\
                Matching with pre-trained transition embedding   & \textbf{28.06\%} & \textbf{66.85\%} & \textbf{5.480}  \\
                \hline
            \end{tabular}
            }
        \end{table}
        \setlength{\tabcolsep}{1.4pt}
        
        \smallskip
        \noindent\textbf{Comparing with the classification method.}
        In this ablation study, we remove the transition embedding, replace triplet margin loss with cross-entropy, and train the recommendation model utilizing the transition category label.
        The comparison result is shown in Table \ref{tab:classification}.
        The classification model performs worse compared with the retrieval model. The reason is that the learned transition embeddings in retrieval model contain richer visual properties of the transitions compared with semantically meaningless one-hot vectors used in the classification model. In addition, the cross-entropy loss may impose excessive punishment for negative categories due to using one-hot ground-truth, thus 
        neglecting the fact that there are similar transitions as the ground-truth transition and they can also be used as favorable alternatives.

\subsection{User Study}
\label{section:user_study}

Since the transition recommendation is subjective, the viewer’s feeling is essential to the evaluation. Therefore, we conduct a user study to further verify its effectiveness. Specifically, we collect raw video shots from online copyright free video sources, e.g. videvo.net\footnote{https://www.videvo.net/}, covering various topics such as travel, life, entertainment, sports, nature, and animals. For each set of video shots, we fix their orders and assign an appropriate background music, leaving only the transitions between neighboring video shots to be added. After selecting 
video transitions, we use the tool of \textit{CapCut} to connect the raw video shots by transitions, producing the final videos.
We compare among following three methods of selecting video transitions. 

\begin{enumerate}
    \item \textbf{Weighted random pick.} At each transition point, select the transition category by a random sampling from a multinomial distribution. The probability of each category is its frequency in our collected video transition dataset.
    \item \textbf{Professional video editor.} We ask a professional editor who has 6 years of video editing to select transitions. He is free to take as long as he wants to select the best transitions depending on his understanding of the given video/audio.
    \item \textbf{Our method.} The top-1 video transition predicted by our method at each transition point is used as the best selection.
\end{enumerate}

\begin{figure}[tbp]
    \centering
    \includegraphics[width=0.8\textwidth]{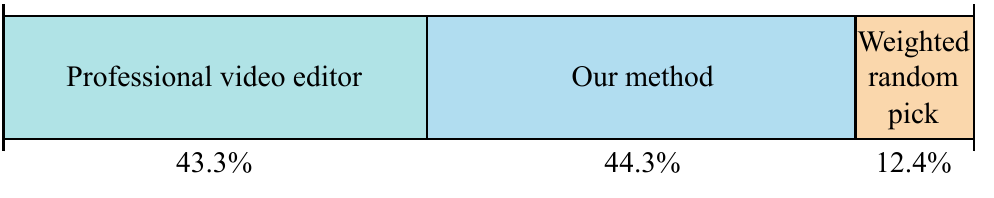}
    \caption{The voting results for three methods in Q1.}
    \label{fig:vote}
\end{figure}

We collect 20 groups of video results in total for user study. Each group contains three videos edited using above three methods respectively. We invite overall 15 non-expert volunteers to participate in the evaluation. 
They are asked to choose a favorite video from each group (Q1) and rate each video on a scale of 1 to 5 (1 = poor, 5 = excellent, Q2), taking into account the general visual quality of videos and the matching degree between transitions and video/audio.
Table \ref{tab:user_study_results} and Fig. \ref{fig:vote} show the statistics of the results. As shown in Table \ref{tab:user_study_results}, the videos generated using random pick receive the lowest score. Our method is pretty close to the professional editing in terms of average score, but being much more efficient by drastically reducing the average processing time of each video from 7.5 minutes to only 1.5 seconds (a \textbf{300\scalebox{1.25}{$\times$}} speedup). Interestingly, in Fig. \ref{fig:vote} which shows the voting results of Q1, one can see that videos from our methods are slightly more appealing than that from the professional editor. Above experimental results clearly demonstrate the advantages of our method in producing high-quality video transitions recommendations.

\setlength{\tabcolsep}{4pt}
\begin{table}[tbp]
    \caption{The statistical results of user study. The inference time is reported as the time cost of our method.}
    \label{tab:user_study_results}
    \centering
    \begin{tabular}{l|cc}
        \hline
        Method & Avg. score & Avg. time (per video) \\
        \hline
        Weighted random pick & 2.96 & - \\
        Professional video editor & 3.80 & 7.5 min \\
        Our method & 3.76 & 1.5 secs \\
        \hline
    \end{tabular}
\end{table}
\setlength{\tabcolsep}{1.4pt}

\section{Conclusions and Future Works}
    
    The recent development of online video tools and platforms creates a high demand for a user-friendly video editing experience, which asks for a computational method or artificial intelligent model to lower the barrier, improve efficiency and ensure quality for doing video editing. Therefore, we propose a new task of video transition recommendation (VTR) to automatically recommend transitions based on any visual and audio inputs. We start with building a large-scale transition dataset. Then we formulate VTR as a multi-modal retrieval problem and propose a flexible framework for addressing the task. Through extensive qualitative and quantitative evaluations, we clearly demonstrate the effectiveness of our method. We hope this work can inspire more researchers to work on VTR and bring creativity and convenience to both professionals and non-professional users. Future works include but are not limited to extending the framework to support more video editing effects 
   like video animation, 3D movements, etc., developing more efficient models for mobile deployment and integrating with other video editing techniques to create more comprehensive video editing systems.
    
\section{Acknowledgement}

Yaojie Shen did this work when interning at ByteDance Inc.
    
\clearpage

\bibliographystyle{splncs}
\bibliography{autotrans}

\appendix
\renewcommand\thefigure{\thesection.\arabic{figure}}
\setcounter{figure}{0} 
\renewcommand{\thetable}{\thesection.\arabic{table}}
\setcounter{table}{0}

\clearpage

\section{Supplementary Material}

In this supplementary material, we provide following ablation studies and analysis. (1) More detailed analysis of the collected dataset. (2) Ablation on freezing part of SlowFast and varying the model size. (3) Analysis of recommendation results. We show that they are reasonable and plausible by following common video editing guidelines. (4) Experimental results of adding ``direct cut'' in transitions recommendation. (5) Visualization of demo videos rendered with 30 transition types.

\subsection{Analysis of Dataset}

\begin{figure}[bp]
    \centering
    \includegraphics[width=\linewidth]{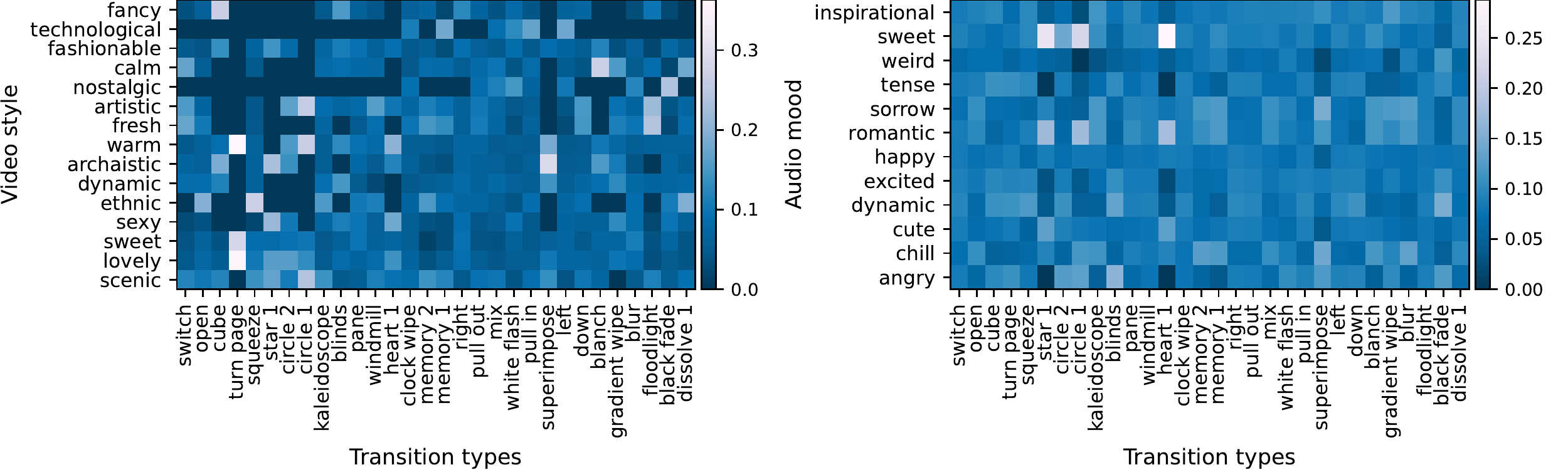}
    \caption{The relationships between video styles, audio mood and transition types. }
    \label{fig:statistic}
\end{figure}

We conduct experiments to analyze more relations between transitions and the properties of visual/audio inputs. Specifically, we use two classifiers of video style/audio mood trained on in-house datasets to get labels of video shots and audio respectively. Then we get the statistical results of the frequencies of transitions w.r.t. each category, followed by a column-wise normalization. As visualized in Fig.~\ref{fig:statistic}, one can get some hints of how transitions match with visual/audio semantics. To give a few examples, ``floodlight'' and ``black fade'' appear more with the visual style of ``fresh'' and ``nostalgic'' respectively. Both ``star'' and ``heart'' tend to come with the audio mood of ``sweet'' and ``romantic''. Note these results comply with general preferences in video editing, which evidences the feasibility of learning meaningful correspondence from inputs to transitions using the dataset. 

\subsection{Model Architecture Ablation}

\noindent\textbf{Impact of freezing SlowFast backbone.}
When training the recommendation model, we freeze stage one to stage three of the SlowFast network by default. To verify its impact, we do some experiments with different freezing proportions of SlowFast network in the setting with only visual as inputs. From the results in Table \ref{tab:freezing}, we can see that freezing all parameters severely damages the performance of the model than freezing the parameters from stage 1-3. While not freezing any parameters only improves performance by a small amount than freezing the parameters from stage 1-3. So in order to balance the performance and efficiency of the algorithm in training, we choose to freeze stage 1-3 of the SlowFast network by default.

\setlength{\tabcolsep}{4pt}
\begin{table}[tbp]
    \caption{The impact of freezing the SlowFast backbone in visual-only setting. By default, we freeze stage 1-3 to facilitate testing.}
    \label{tab:freezing}
    \centering
    \begin{tabular}{l|ccc}
        \hline
        SlowFast Freezing       & Recall@1 & Recall@5 & Mean Rank \\
        \hline
        Freeze all parameters   & 22.39\%& 26.53\% & 6.097 \\
        Freeze stage 1-3        & 25.40\% & 66.33\% & 5.665 \\
        No freezing             & \textbf{25.97\%} & \textbf{66.95\%} & \textbf{5.579} \\
        \hline
    \end{tabular}
\end{table}
\setlength{\tabcolsep}{1.4pt}

\smallskip
\noindent\textbf{Variations on the model architecture.}
In this experiment, we study the influence of the different model architecture and the embedding dimension. In Table \ref{tab:model_size} (a), the influence of different dimension in the embedding matching space is investigated. Reducing the dimension of matching embedding does harm to the performance. 
From Table \ref{tab:model_size} (b) and (c), we observe that a smaller feature dimension or number of the transformer layers in the transformer network also hurts performance. At the same time, a larger feature dimension or number of the transformer layers has adverse effects on performance, indicating that large model size may cause severe over-fitting.

\setlength{\tabcolsep}{4pt}
\begin{table}[tbp]
\begin{center}
    \caption{Ablation study on the model size. Where $N$ is the number of transformer encoder layers, $d_\mathrm{model}$ is the dimension of transformer layers, and $d_\mathrm{matching}$ is the dimension of the common matching space.}
    \label{tab:model_size}
    \begin{tabular}{c|ccc|ccc}
        \hline
        &$N$  &  $d_\mathrm{model}$  & $d_\mathrm{matching}$ & Recall@1 & Recall@5 & Mean Rank \\
        \hline
        base & 2 & 2048 & 2048 & \textbf{28.06\%} & \textbf{66.85\%} & \textbf{5.480} \\
        \hline
        \multirow{2}{*}{(a)}&2 & 2048 & 1024 & 26.59\% & 67.09\% & 5.493 \\
        &2 & 2048 & 512  & 26.40\% & 67.07\% & 5.499 \\
        \hline
        \multirow{2}{*}{(b)} &2 & 1024 & 2048 & 25.52\% & 66.71\% & 5.598\\
        &2 & 4096 & 2048 & 26.93\% & 66.55\% & 5.541 \\
        \hline
        \multirow{2}{*}{(c)} & 1 & 2048 & 2048 & 25.77\% & 66.45\% & 5.623 \\
        &4 & 2048 & 2048 & 27.26\% & 66.47\% & 5.528 \\
        \hline
    \end{tabular}
\end{center}
\end{table}
\setlength{\tabcolsep}{1.4pt}

\subsection{Results Analysis}

In this section, we show through concrete examples that our method indeed learns the general guidelines of using transitions in video editing. Therefore, the videos generated by our method comply with common cinematography knowledge and aesthetic principles.

\begin{figure}[tp]
    \centering
    \includegraphics[width=0.99\textwidth]{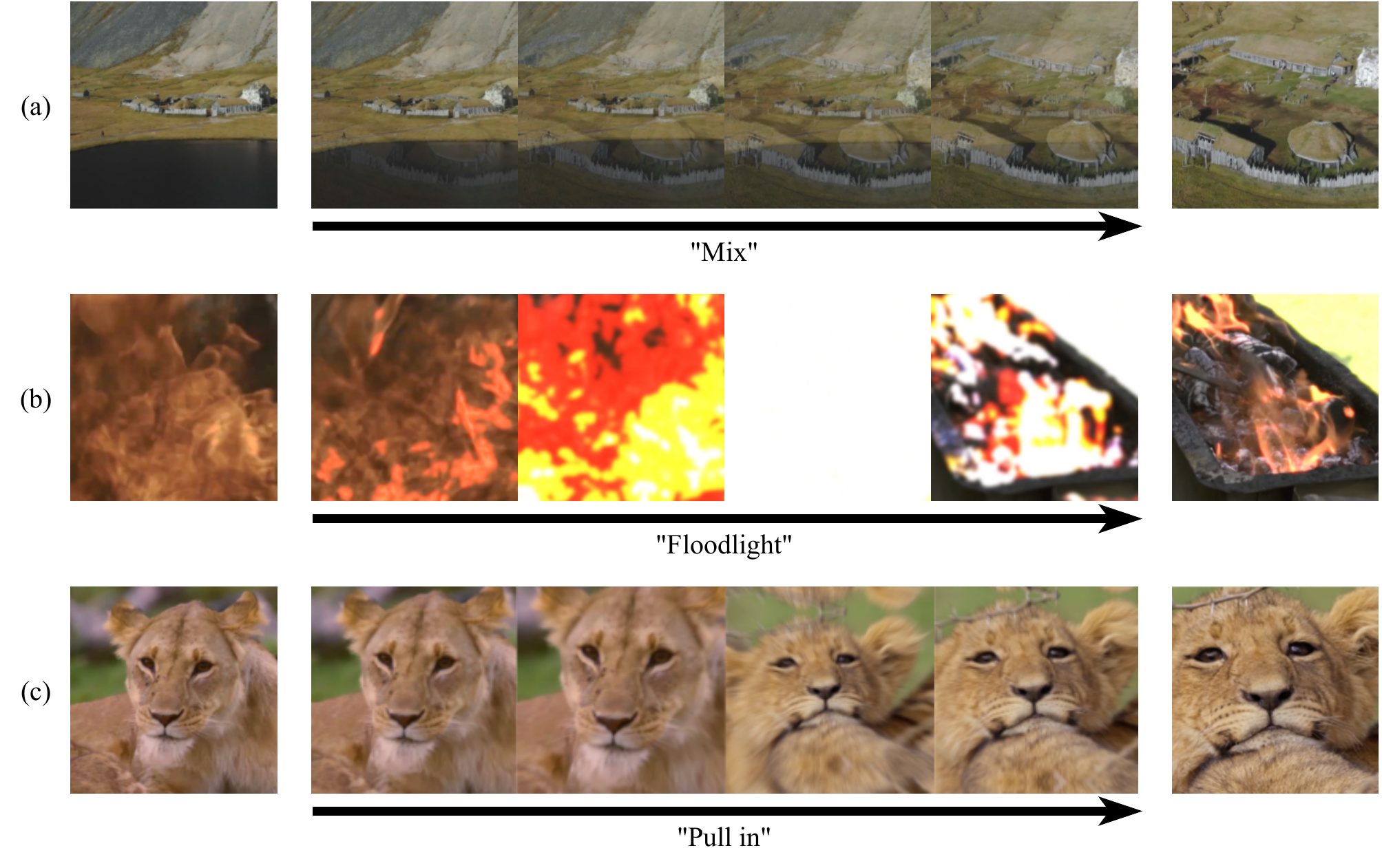}
    \caption{The video transition recommendations generated by our method. In each row, the leftmost and the rightmost images represents the video shots before and after transitions respectively. The frames in middle represent the video clip in transition. For better visualization, please refer to videos (a)-(c) under folder ``video/fig-A-2''.}
    \label{fig:transition_sample}
\end{figure}

As shown in Fig. \ref{fig:transition_sample}, our method can capture the visual changes in neighboring video shots and thus recommend suitable video transitions. In Fig. \ref{fig:transition_sample} (a), since the neighboring video shots have similar scenes, our method ranks gentle transitions higher, such as ``mix", ``black fade", ``dissolve" and ``blur", in order to keep the coherence of video narration.
In Fig. \ref{fig:transition_sample} (b) where neighboring video shots have large scene changes and brightness changes, our method recommends more dynamic transitions like ``floodlight". 
In Fig. \ref{fig:transition_sample} (c) where a long video shot switches to a shorter one, our method recommends the ``pull in" to ensure natural transition .

\begin{figure}[tp]
    \centering
    \includegraphics[width=0.99\textwidth]{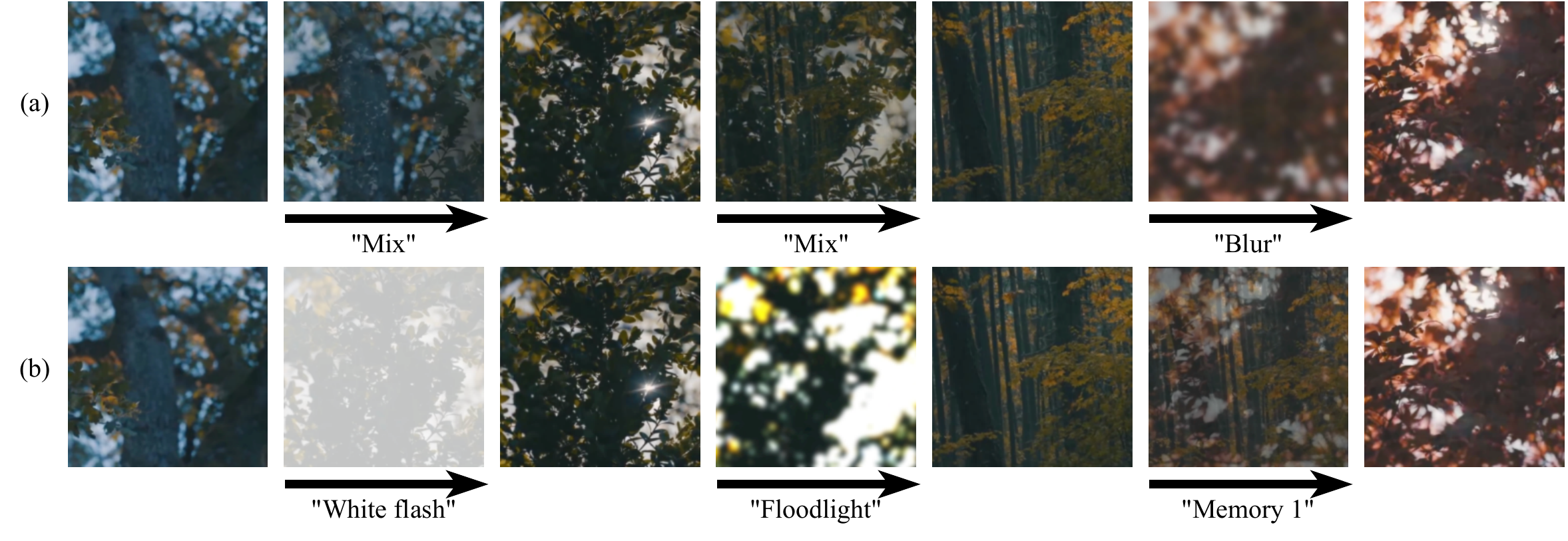}
    \caption{Transitions recommended for video shots with soft background music (see videos (a)-(b) under folder ``video/fig-A-3''). (a) and (b) corresponds to the results of our method and weighted random pick respectively.}
    \label{fig:seq_sample}
\end{figure}

\begin{figure}[tp]
    \centering
    \includegraphics[width=0.99\textwidth]{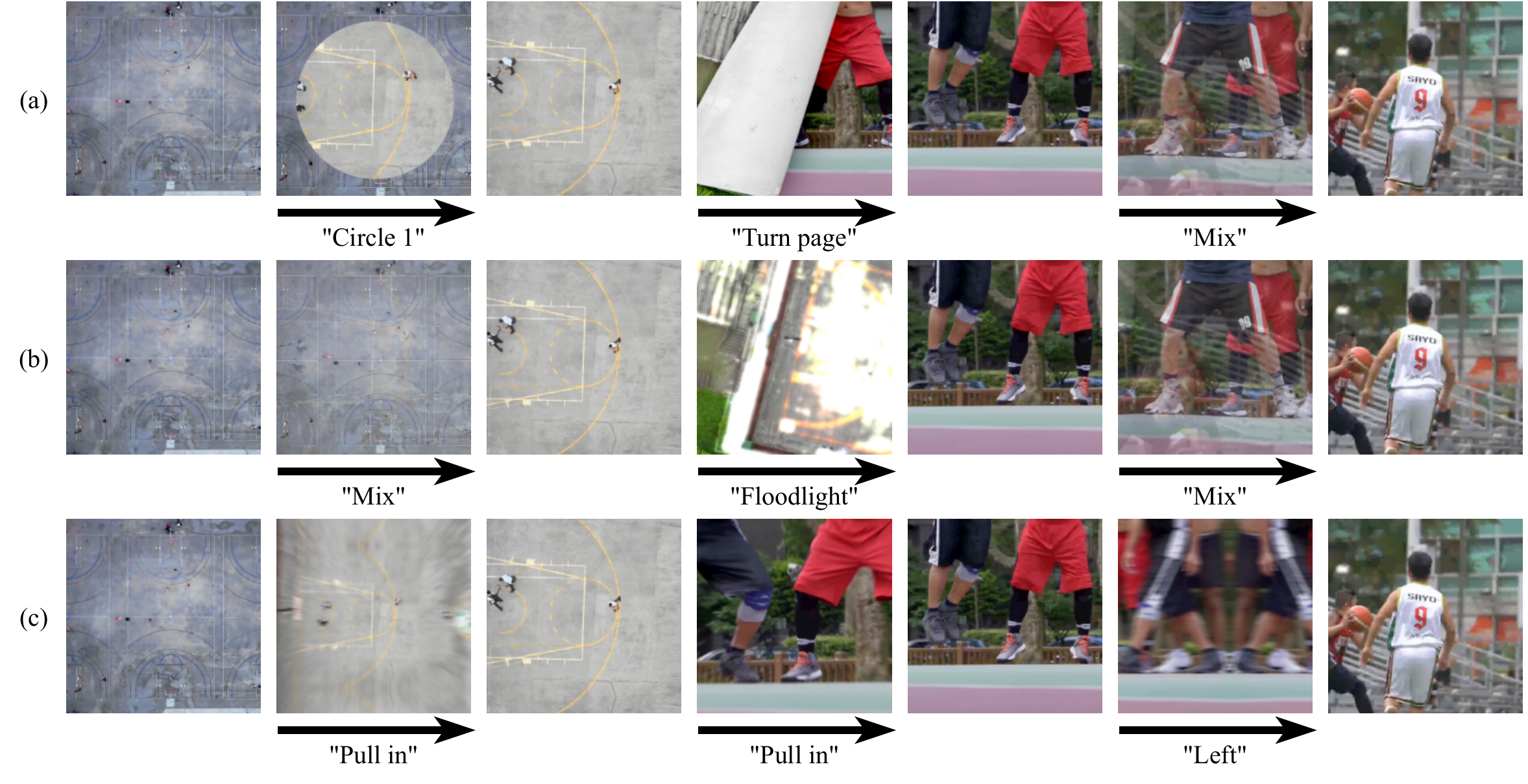}
    \caption{A comparison of the transitions selected by three different methods (see videos (a)-(c) under folder ``video/fig-A-4''). Transitions in (a), (b) and (c) are selected by weighted random pick, our method and professional editor respectively.}
    \label{fig:seq_sample_2}
\end{figure}

In Fig. \ref{fig:seq_sample} (a), we show an example that our method is able to adapt to the characteristics of music. Since the background music used in this example is with soft tune, our method prefers to recommend gentle transitions. Otherwise using abrupt transitions may break the visual-auditory harmonious. In contrast, 
since the weighted random pick method does not consider visual/audio contents, its results are less reasonable (e.g. using too much ``flashing'') as shown in Fig. \ref{fig:seq_sample} (b).

We show another comparing example among three methods in Fig. \ref{fig:seq_sample_2}. Compared with weighted random pick (Fig. \ref{fig:seq_sample_2} (a)), our method (Fig. \ref{fig:seq_sample_2} (b)) recommends better results which are reflected by the consistency in sequential transitions predictions, as well as the nice matching with visual and audio contents. While the professional editor (Fig. \ref{fig:seq_sample_2} (c)) may be advantageous 
in capturing the details in videos like the specific movement pattern of basketball players to select dynamic transitions, we note our method is much faster in terms of efficiency. This comparison result also motivates us to use more fine-grained video embedding to further improve our method, which we leave for future work.

\subsection{``Direct Cut'' and More User Study}

\begin{figure}[tp]
    \centering
    \includegraphics[width=0.5\textwidth]{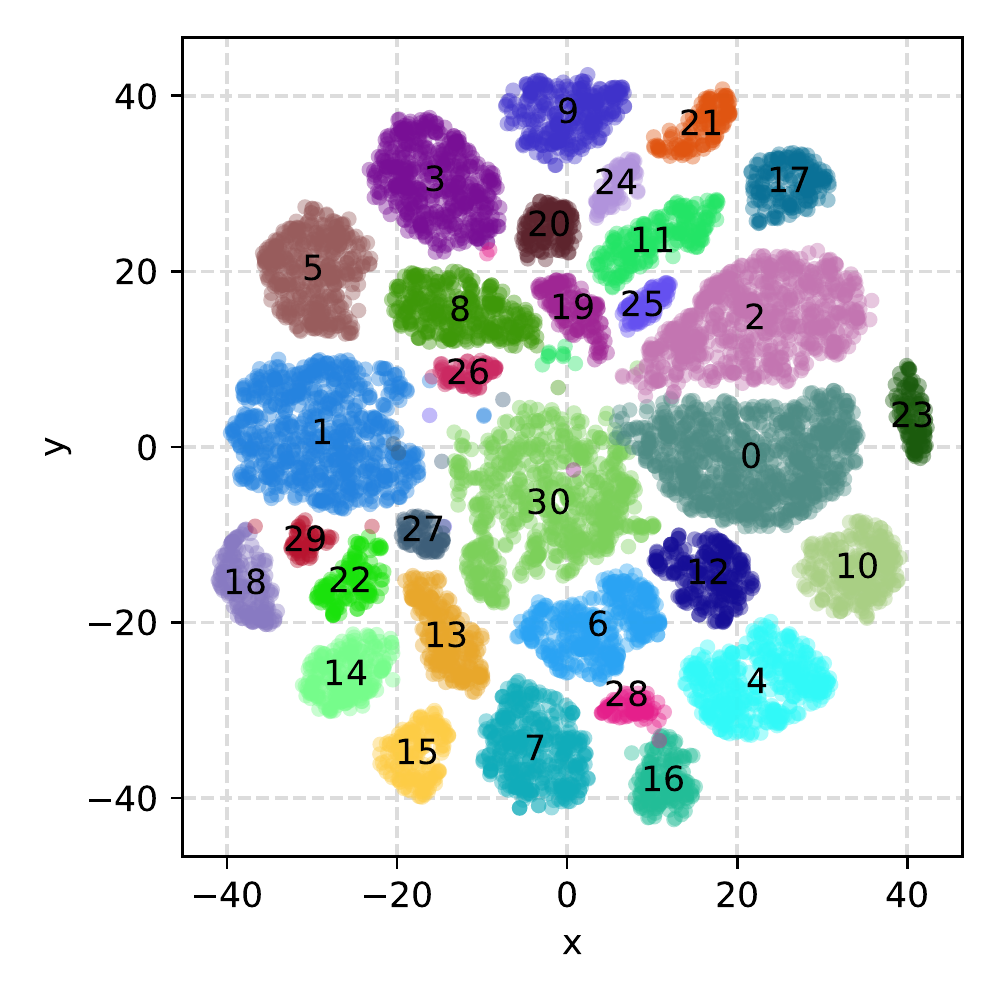}
    \caption{t-SNE visualization of the pre-trained transition embedding  (with ``direct cut'', the index of ``direct cut'' is 30).}
    \label{fig:tsne_with_direct_cut}
\end{figure}

We note that ``direct cut'' is also widely used in video editing by directly connecting shots without using any effects. We add ``direct cut'' and re-train/re-evaluate our model to verify the extendibility of our proposed framework.
We draw similar conclusions from this new experiment as in the main paper and verify both quantitatively and qualitatively that our method can handle this case. Fig.~\ref{fig:tsne_with_direct_cut} shows the learned embeddings of transitions including ``direct cut''. The semantic relationships are still preserved. As shown in Table~\ref{tab:exp}, the performance of model with ``direct cut'' is on par with that without ``direct cut'', showing the model can successfully learn the matching from input to transitions.

\begin{table}[tp]
    \caption{Ablative experiments on direct cut.}
    \label{tab:exp}
    \centering
    \begin{tabular}{l|c|c|ccc}
        \hline
        Modal & with ``direct cut'' & R@1 $\uparrow$ & R@5 $\uparrow$ & Mean Rank $\downarrow$ \\
        \hline
        Visual+Audio &  & 28.06\%& 66.85\%& 5.480 \\
        Visual+Audio&\checkmark & 30.57\%& 67.98\% & 5.347 \\
        \hline
    \end{tabular}
\end{table}

\begin{figure}[tp]
\centering
\includegraphics[width=0.95\linewidth]{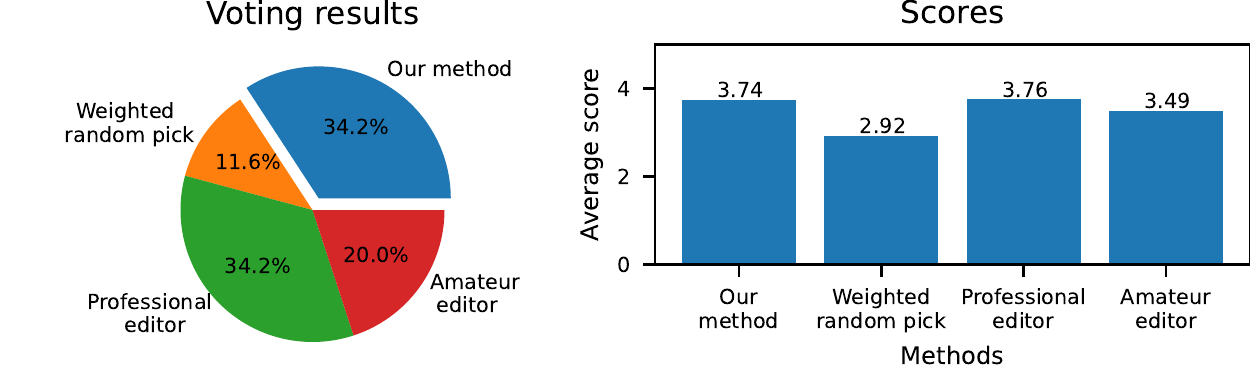}
\caption{User study results with more editors, ``direct cut'' is allowed in all comparing methods.}
\label{fig:user_study}
\end{figure}

To further verify the effectiveness of our method, we conduct a more comprehensive user study following the practice introduced in Section \ref{section:user_study} in the main paper. In this experiment, we hire 10 professional and 10 amateur editors, and direct cut is allowed to use in all comparing methods. As show in Fig.~\ref{fig:user_study}, in terms of both voting result and average score, our method surpasses amateur editors and achieves comparable results with professional editors, therefore demonstrating the effectiveness of our method.

\subsection{Example videos of all transition effects}
To help readers understand video transitions more straightforwardly, we provide example videos created using video transitions and a fixed pair of video shots. Overall 30 demo videos are included in the zipped file, under the folder ``video/transitions''.

\end{document}